\documentclass[runningheads]{llncs}
\usepackage{graphicx}
\usepackage{booktabs}

\usepackage{amsmath,amssymb} 
\usepackage{color}
\usepackage{amsmath}
\usepackage{tikz}
\usepackage{multirow}
\usepackage{algorithmic}
\usepackage{algorithm}

\begin{document}
\newcommand*\samethanks[1][\value{footnote}]{\footnotemark[#1]}
\pagestyle{headings}
\mainmatter
\def\ECCV16SubNumber{4\textbf{}}  

\title{Noisy Student Training using Body Language Dataset Improves Facial Expression Recognition} 
\titlerunning{Noisy Student Training Improves Facial Expression Recognition}
%
\author{Vikas Kumar\inst{\thanks{equal contribution}}
\and
Shivansh Rao\inst{\samethanks}
\and
Li Yu\inst{}
}
\authorrunning{V. Kumar et al.}
%
\institute{
The Pennsylvania State University, University Park
\email{\{vuk160,sqr5687,luy133\}@psu.edu}
}

\maketitle

\begin{abstract}
Facial expression recognition from videos in the wild is a  challenging task due to the lack of abundant labelled training data. Large DNN (deep neural network) architectures and ensemble methods have resulted in better performance, but soon reach saturation at some point due to data inadequacy. In this paper, we use a self-training method that utilizes a combination of a labelled dataset and an unlabelled dataset (Body Language Dataset - BoLD). Experimental analysis shows that training a noisy student network iteratively helps in achieving significantly better results. Additionally, our model isolates different regions of the face and processes them independently using a multi-level attention mechanism which further boosts the performance.  Our results show that the proposed method achieves state-of-the-art performance on benchmark datasets CK+ and AFEW 8.0 when compared to other single models. Code available at \textcolor{blue}{github.com/vkumar1997/Emotion-BEEU}

\end{abstract}

\section{Introduction}
Automatic facial expression recognition from images/videos has many applications such as human-computer interaction (HCI), bodily expressed emotions, human behaviour understanding, and has thus gained a lot of attention in academia and industry. Although there has been extensive research on this subject, facial expression recognition in the wild remains a challenging problem because of several factors such as occlusion, illumination, motion blur, subject-specific facial variations, along with the lack of extensive labelled training datasets. Following a similar line of research, our task aims to classify a given video in the wild to one of the seven broad categorical emotions. We propose an efficient model that addresses the challenges posed by videos in the wild while tackling the issue of labelled data inadequacy. The input data used for facial expression recognition can be multi-modal, i.e. it may have visual information as well as audio information. However, the scope of this paper is limited to emotion classification using only visual information.

Most of the recent research on the publicly-available AFEW 8.0 (Acted Facial Expressions in the Wild) \cite{dhall2012collecting} dataset has focused on improving accuracy without regard to computational complexity, architectural complexity, energy \& policy considerations, generality, and training efficiency. Several state-of-the-art methods \cite{fan2018video,lu2018multiple,vielzeuf2017temporal} on this dataset have originated from the EmotiW \cite{dhall2019emotiw} challenge with no clear computational-cost analysis. Fan et al. \cite{fan2018video} achieved the highest validation accuracy based on visual cues, but they used a fusion of five different architectures with more than 300 million parameters. In contrast, our proposed method uses a single model with approximately 25 million parameters and comparable performance.

While previous work focused on improving performance by increasing model capacity, our method focuses on better pre-processing, feature selection, and adequate training. Prior research \cite{littlewort2004dynamics,shan2009facial,knyazev2017convolutional,tang2013deep} uses simple aggregation or averaging operation on features from multiple frames to form a fixed-dimensional feature vector. However, such methods do not account for the fact that a few principal frames in a video can be used to identify the target emotion, while the rest of the frames have a negligible contribution. Frame-attention has been used \cite{meng2019frame} for selectively processing frames in a video, but it can further be coupled with spatial-attention which could identify the most discriminative regions in a particular frame. We use a three-level attention mechanism in our model: a) spatial-attention block that helps to selectively process feature maps of a frame, b) channel-attention block that focuses on the face regions at a local and a global level, i.e. eyes region (upper face), mouth region (lower face) and whole face, and c) frame-attention block that helps to identify the most important frames in a video.

AFEW 8.0 \cite{dhall2012collecting} has several limitations (Sec. \ref{sec: related}) that restricts the generalization capabilities of deep learning models. To overcome these limitations, we use an unlabelled subset of the BoLD dataset \cite{luo2020arbee} for semi-supervised learning. Inspired by Xie et al. \cite{xie2019self}, we use a teacher-student learning method where the training process is iterated by using the same student again as the teacher. During the training of the student, noise is injected into the student model to force it to generalize better than the teacher. Results show that the student performs better with each iteration, hence improving the overall accuracy on the validation set.

The rest of the paper is organized as follows. Sec. \ref{sec: dataset} explains the datasets (AFEW 8.0 \cite{dhall2012collecting}, CK+ \cite{lucey2010extended} and BoLD \cite{luo2020arbee}) that are used for training our model along with the pre-processing pipeline used for face detection, alignment and illumination correction. Sec. \ref{sec: backbone} explains the backbone network and covers the three types of attention and its importance in detail. Sec. \ref{sec: student} covers the use of the BoLD dataset for iterative training and the experimental results of semi-supervised learning. Sec. \ref{sec: compare} compares the results of our methods to other state-of-the-art methods on the AFEW 8.0 dataset. Additionally, we use another benchmark dataset CK+ \cite{lucey2010extended} (posed conditions) as well as perform ablation studies (Sec. \ref{sec: ablation}) to prove the validity of our model and training procedure.

\section{Related Work} \label{sec: related}
\textbf{Facial Expression Recognition:} A number of methods have been proposed on the AFEW 8.0 dataset \cite{dhall2012collecting} since the first EmotiW \cite{dhall2019emotiw} challenge in 2013. Earlier approaches include non-deep learning methods such as multiple kernel learning \cite{sikka2013multiple}, least-square regression on grassmanian manifold \cite{liu2013partial}, and feature fusion with kernel learning \cite{chen2014emotion}, whereas recent approaches include deep-learning methods such as frame-attention networks \cite{meng2019frame}, multiple spatial-temporal learning \cite{lu2018multiple}, and deeply supervised emotion recognition \cite{fan2018video}. Although several methods \cite{fan2018video,lu2018multiple,vielzeuf2017temporal,liu2018multi} have achieved impressive results on the AFEW 8.0 dataset, many have used ensemble (fusion) based methods and considered multiple modalities without commenting on the resources and time required to train such models. 
Spatial-temporal methods \cite{vielzeuf2017temporal,fan2016video} aim to model motion information or temporal coherency in the videos  using 3D Convolution \cite{tran2015learning} or LSTM (Long short-term memory) \cite{hochreiter1997long}. However, owing to computational efficiency and the ability to treat sequential information with a global context, several studies \cite{meng2019frame,aminbeidokhti2019emotion} related to facial expression recognition have successfully implemented attention-based methods by assigning a weight to each timestep in the video. Similarly, spatial self-attention has been used \cite{aminbeidokhti2019emotion,fang2019self,lin2017structured} as a means to guide the process of feature extraction and find the importance of each local image feature. Our model builds upon the spatial self-attention mechanism and additionally uses a channel-attention mechanism to exploit the differential effects of facial feedback signals from the upper-face and lower-face regions \cite{wang2020region,zeng2018false}.

\noindent\textbf{Training Datasets:} Despite being a long-established dataset, AFEW 8.0 \cite{dhall2012collecting} has several shortcomings. Firstly, the dataset contains significantly fewer training examples for fear, surprise and disgust categories which makes the dataset imbalanced. Secondly, the videos are extracted from mainstream cinema, and scenes depicting fear are often shot in the dark, which again makes the model biased towards other categories  \cite{lu2018multiple,acharya2018covariance}. Such limitations warrant the use of additional datasets for better generalization. However, not many in-the-wild labelled video datasets are publicly available for facial expression recognition. Several related datasets \cite{lucey2010extended,valstar2010induced,lyons1998japanese} are captured in posed conditions and are restricted to a certain country or community. Aff-Wild2 \cite{kollias2018aff} is another popular dataset, but it contains per-frame annotations, and thus cannot be used in our work which performs video-level classification based on facial expressions. We use an unlabelled portion of the BoLD dataset \cite{luo2020arbee} since the videos are of the desired length and are captured from movies similar to our labelled dataset. 

\noindent\textbf{Semi-Supervised Learning:} The semi-supervised approach is effective in classification problems when the labelled training data is not sufficient. We use noisy student training \cite{xie2019self} for semi-supervised learning, in which the trick involves the student to be deliberately noised when it trains on the combined labelled and unlabelled dataset. Input noise is added to the student model in the form of data augmentations, which ensures that different alterations of the same video should have the same emotion, hence making the student model more robust. Additionally, model noise is added in the form of dropout, which forces the student (single model) to match the performance of an ensemble model. Other techniques for semi-supervised learning include self-training \cite{yarowsky1995unsupervised,riloff1996automatically}, data-distillation \cite{radosavovic2018data} and consistency training \cite{bachman2014learning,rasmus2015semi}. Self-training is similar to noisy student training, but it does not use or justify the role of noise in training a powerful student. Data-distillation uses the approach of strengthening the teacher using ensembles instead of weakening the student; however, a smaller student makes it difficult to mimic the teacher. Consistency training adds regularization parameters to the teacher model during training to induce invariance to input and model noise, resulting in confident pseudo-labels. However, such constraints lead to lower accuracy and a less powerful teacher \cite{xie2019self}.

\begin{figure}[t!]
\centering
\includegraphics[width=1.0\textwidth]{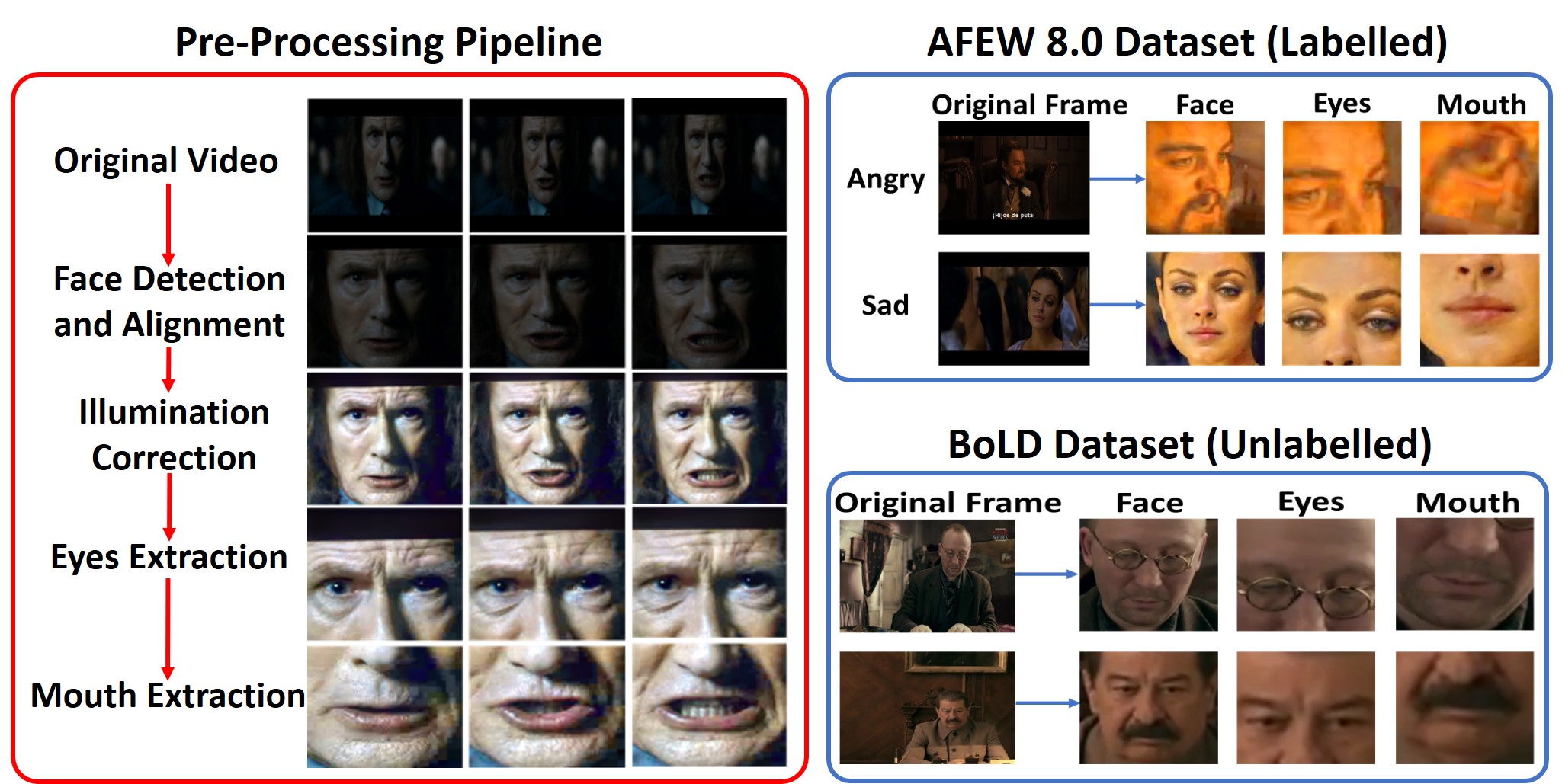}
\caption{The pre-processing steps mainly include face detection and alignment (MTCNN \cite{zhang2016joint}), illumination correction (Enlighten-GAN \cite{jiang2019enlightengan}) and landmark-based cropping. Examples from labelled dataset (AFEW 8.0) and unlabelled dataset (BoLD dataset) are shown. As seen in the figure, only videos with a close shot of the face are selected from the BoLD dataset.}
\label{fig:dataset}
\end{figure}

\section{Dataset} \label{sec: dataset}
In this section, we first describe the datasets that we use in our experiments, followed by the pre-processing pipeline. \newline

\noindent\textbf{Labelled Sets:} AFEW 8.0 (Acted Facial Expression in the Wild) \cite{dhall2012collecting} contains videos with seven emotion labels, i.e. anger (197 samples), neutral (207 samples), sad (179 samples), fear (127 samples), surprise (120 samples), happiness (212 samples), and disgust (114 samples) from different movies. The train set consists of 773 video samples (46,080 frames), and the validation set consists of 383 video samples (21,157 frames). The results are reported on the validation set since the test set labels are only available to EmotiW challenge \cite{dhall2019emotiw} participants. Some of the example frames are shown in Fig. \ref{fig:dataset}. CK+ (Cohn Kanade Extended) \cite{lucey2010extended} contains 327 video sequences (5878 frames) divided into seven categories, i.e anger (45 samples), disgust (59 samples), fear (25 samples), happy (69 samples), sad (28 samples), surprise (83 samples), and contempt (18 samples). The motivation behind testing our method on a posed dataset is to establish the robustness of our model and semi-supervised learning method irrespective of the data source. Since CK+ does not have a testing set, we report the average accuracy obtained using 10-fold cross-validation as seen in other studies \cite{meng2019frame,zhang2017facial,jung2015joint,cai2018island,sikka2016lomo}. \newline

\noindent\textbf{Unlabelled Set:} BoLD (Body Language Dataset) \cite{luo2020arbee} contains videos selected from the publicly available AVA dataset \cite{gu2018ava}, which contains a list of YouTube movie IDs. While the gathered videos are annotated based on body language, the videos having a close shot of the face instead of the whole or partially-occluded body are unlabelled. To create an AFEW-like subset from the BoLD dataset, we impose two conditions to automatically validate a video. Firstly, a video should have $f\,(\geq 30)$ such consecutive frames where only one actor's face is detected by MTCNN (Multi-task Cascaded Convolutional Networks) \cite{zhang2016joint}. Secondly, the bounding box of the face detected using MTCNN should exceed an occupied area threshold for the majority of those $f$ frames. If the video satisfies the above two conditions, a smaller video with those $f$ frames is added to the unlabelled dataset. Using this procedure, we create a subset of 3450 videos (224,258 frames) from the original BoLD dataset. Some of the examples gathered are shown in Fig. \ref{fig:dataset}. \newline

\noindent\textbf{Pre-Processing:} \label{sec: preprocess}
Previous work \cite{lu2018multiple,meng2019frame} have used CNN-based detector provided by dlib \cite{king2009dlib} for face alignment. However, the alignment of faces is highly dependent on accurate detection of facial landmarks and CNN-based detector provided by dlib is not reliable for `in-the-wild' conditions (especially non-frontal faces). We use MTCNN \cite{zhang2016joint} for face detection and alignment. If MTCNN detects multiple faces in a frame, the face with the largest bounding box is selected. After obtaining the facial landmarks, its alignment is corrected using the angle between the line connecting the landmark points of the eyes and the horizontal line. After detection and alignment, the cropped face is resized to 224*224 pixels, which is the input size of our model. 

We use the landmarks given by MTCNN to isolate the mouth (lower face) and eyes (upper face) region. The upper face is isolated using the eyes landmarks with the desired left eye normalized co-ordinates being (0.2, 0.6) and right-eye co-ordinates being (0.8, 0.6) in the new frame, which is enough to occlude the lower-half of the face in almost all frames (Fig. \ref{fig:dataset}). A similar procedure is used for occluding the upper-half of the face and isolating the mouth region using left-mouth and right-mouth landmarks. All landmark-based crops are again resized to 224*224 pixels.

As addressed earlier, some of the categories of emotions are often captured in the dark in movies, which requires an illumination correction step. Several methods have been suggested for illumination normalization such as gamma correction \cite{anila2012preprocessing,liu2017facial}, Difference of Gaussians (DoG) \cite{wang2012improved} and histogram equalization \cite{bendjillali2019improved,karthigayan2007development} which are effective for facial expression recognition. However, these methods tend to amplify noise, tone distortion, and other artefacts. Hence, we use a state-of-the-art pre-trained deep learning model, i.e. Enlighten-GAN \cite{jiang2019enlightengan} (U-Net \cite{ronneberger2015u} as generator) which provides appropriate results (Fig. \ref{fig:dataset}) with uniform illumination and suppressed noise.

\section{Methodology}
Our proposed methodology is divided into two phases, i.e. a) architecture implementation that defines the backbone network with the three-level attention mechanism, and b) semi-supervised learning.

\begin{figure}[t!]
\centering
\includegraphics[width=1.0\textwidth,height=79mm]{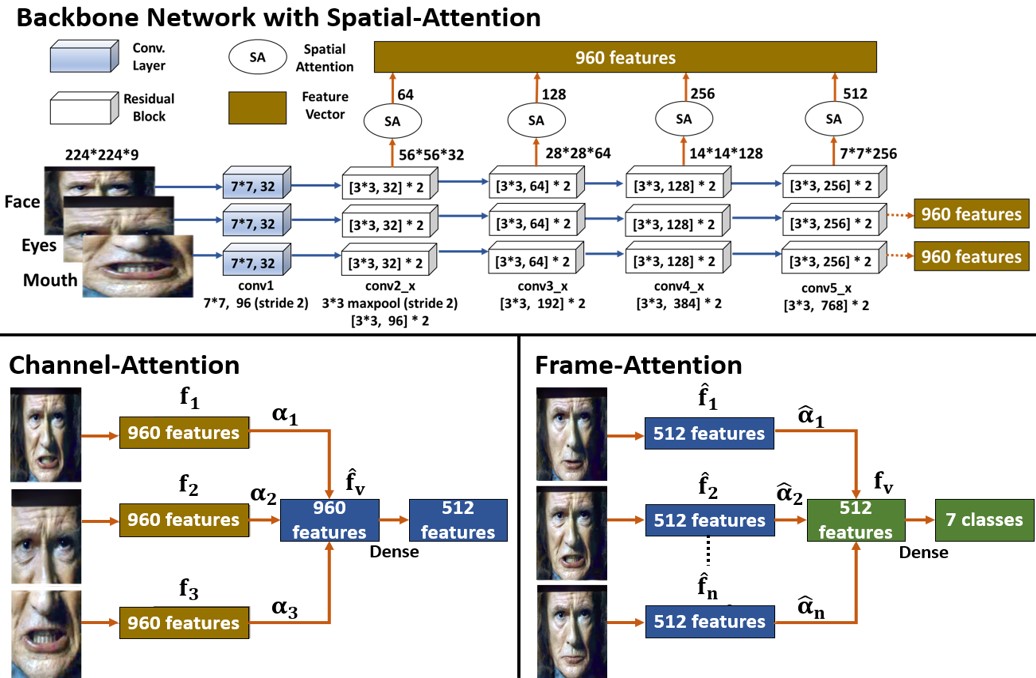}
\caption{Figure shows the backbone network (ResNet-18) and the three-level attention mechanism. Inputs are first processed via Spatial-Attention, followed by Channel-Attention and finally by Frame-Attention.}
\label{fig:model}
\end{figure}

\subsection{Architecture}
\subsubsection{Backbone Network:} \label{sec: backbone}
We use ResNet-18 \cite{he2016deep} architecture as our backbone network, with minor modifications to increase its computational efficiency. Features from each residual block are combined to form the final feature vector (see Fig. \ref{fig:model}). Hence, the final vector has a “multi-level knowledge” from all the residual blocks, ensuring more diverse and robust features. The model is first pre-trained on the FERPlus dataset \cite{barsoum2016training}. Our input at frame-level is an image with nine channels (RGB channels from the face, eyes, and mouth region). To process them independently, the model uses group convolution \cite{krizhevsky2012imagenet} (groups = 3), i.e. it uses a different set of filters for each of the three regions to get the final output feature maps. Group convolution results in a lower computational cost since each kernel filter does not have to convolve on all the feature maps of the previous layer. Simultaneously, it allows data parallelism where each filter group is learning a unique representation and forms a global (face region) or local (eyes and mouth region) context vector from each frame of a video. To allow more filters per group, we increase the number of filters in each residual block, as shown in Fig. \ref{fig:model}.

\begin{figure}[t!]
\centering
\includegraphics[width=1.0\textwidth]{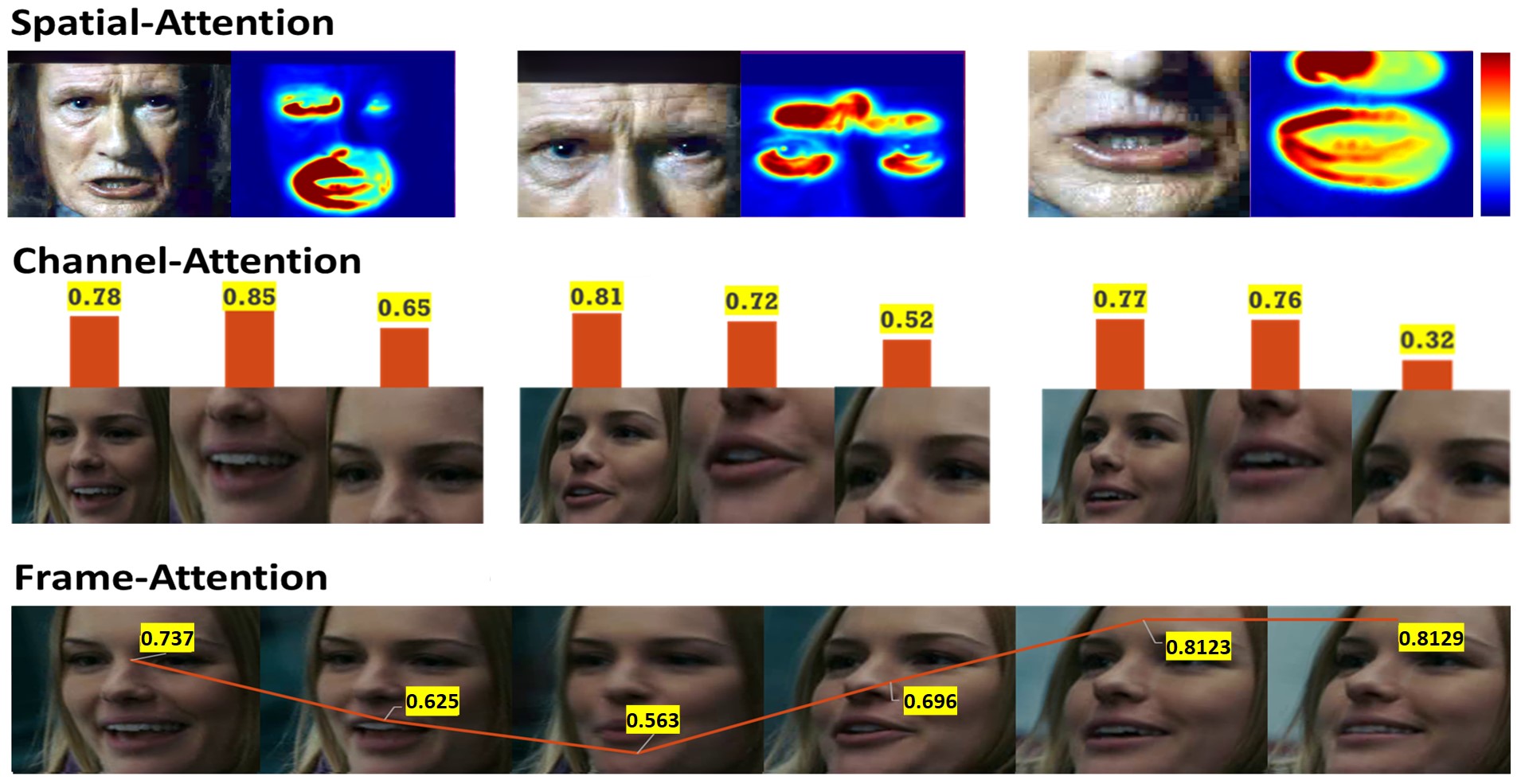}
\caption{This figure shows how multi-level attention works in the proposed method. Spatial-attention (from last residual block) chooses the dominant feature maps from each region. Channel-attention picks the most important region that most clearly shows the target emotion. Frame-attention assigns the salient frames a higher weight.}
\label{fig:attention}
\end{figure}

\subsubsection{Spatial-Attention:} \label{sec: spatial}
A common approach in previous methods is a simple aggregation or average pooling of feature maps to form a fixed dimensional feature vector. However, we use spatial-attention \cite{lin2017structured} that concatenates the feature maps based on the attention weight it has been assigned. Let us assume the output from a residual block is of shape $C=H\times W \times D$ where $H$ and $W$ are the output height and width, and $D$ is the number of output filters. This 3D tensor $C$ is reshaped to a 2D matrix $L$ of shape $R\times D$ where $R = H*W$. The spatial-attention mechanism takes the input matrix $L$ and outputs a weight matrix M of shape $h\times R$ ($h=2$, $h$ is for multiple hops of attention). Each row of the output matrix represents a different hop of attention, and each column has normalized weights due to softmax (see Equation \ref{eq:multi_hop}). The objective is to find the weighted average of R frame descriptors to obtain a vector $v$ of length $D$ (or $h*D$ with multiple hops). 

\begin{gather} 
    M=softmax(W_{s2}tanh(W_{s1}L^T))\label{eq:multi_hop}\\
    v=flatten(M*L)\label{eq:flatten}
\end{gather}

Equation \ref{eq:multi_hop} represents multi-head spatial-attention where $W_{s1}$ is of shape $U \times D$ and $W_{s2}$ is of shape $h \times U$ (U can be set arbitrarily). From this, we obtain flattened vector $v$ using Equation \ref{eq:flatten}. The spatial-attention module is applied on each residual block (see Fig. \ref{fig:model}) and the output vectors are aggregated to obtain a final vector of length $l=960$ each for face ($f_1$), eyes ($f_2$) and mouth ($f_3$) regions. The advantages of spatial attention can be seen in Fig. \ref{fig:attention}. While the feature vector from the face is encoded with a global context, the feature maps from the eyes and mouth region have additional information regarding the minute expressions such as furrowed brow or flared nostrils.

\subsubsection{Channel-Attention:} \label{sec: channel}
Let $f_1$, $f_2$, and $f_3$ be the feature vectors obtained from the face, the eyes, and the mouth region respectively. We model the cross-channel interactions using a lightweight attention module. We use two fully-connected layers to obtain a weight $\alpha$ (Equation \ref{eq:alpha}) for each channel group using which we obtain a weighted average $\hat{f}_v$ (Equation \ref{eq:channel}) of the three feature vectors. ReLU (Rectified Linear Unit) activation is used after the first layer to capture non-linear interactions among the channels.

\begin{gather}
    \alpha_i = \sigma(w^T(ReLU(W^Tf_i))) \label{eq:alpha}\\
    \hat{f}_{v} = \frac{\sum_{i=1}^{3}\alpha_i*f_i}{\sum_{i=1}^{3}\alpha_i}\label{eq:channel}
\end{gather}

where $\sigma$ is the sigmoid activation function, $w$ is a vector of length $r$ (set arbitrarily), and $W$ is a matrix of shape $l\times r$. In Fig. \ref{fig:attention}, we see that the model assigns more weight to the mouth region instead of the eyes region for an expression depicting happiness which is consistent with our findings that mouth region is more prominent for the happy category (Fig. \ref{fig:confusion}).

\subsubsection{Frame-Attention:} \label{sec: frame}
For a video having n frames, we obtain vector $\hat{f}_i$ of length $\hat{l}$ from each frame after the channel-attention module. Finally, we use frame-attention to assign the most discriminative frames a higher weight. Following a similar intuition as in channel-attention, we use two fully-connected layers to obtain a weight $\hat{\alpha}$ (Equation \ref{eq:alphahat}) for each frame using which we find a weighted average $f_v$ (Equation \ref{eq:frame}) of the frame features.

\begin{gather}
    \hat{\alpha_i} = \sigma(\hat{w}^T(ReLU(\hat{W}^T\hat{f_i}))) \label{eq:alphahat}\\
    f_{v} = \frac{\sum_{i=1}^{n}\hat{\alpha_i}*\hat{f_i}}{\sum_{i=1}^{n}\hat{\alpha_i}}\label{eq:frame}
\end{gather}
where $\hat{w}$ is a vector of length $\hat{r}$ (set arbitrarily), and $\hat{W}$ is a matrix of shape $\hat{l}\times \hat{r}$. Fig. \ref{fig:attention} shows how the model assigns a higher weight to the frames which distinctively contains expression depicting happiness. The feature vector $f_v$ is passed through a fully-connected layer to obtain the final 7-dimensional output. 

\subsubsection{Implementation Details:}
We use weighted cross-entropy as our loss function where class weights are assigned based on number of training samples to alleviate the problem of unbalanced data. Additionally, $M$ (Equation \ref{eq:multi_hop}) is regularized by adding the frobenius norm of matrix $MM^T-I$ to the loss function which enforces multi spatial-attention to focus on different regions \cite{lin2017structured}. We use Adam optimizer with an initial learning rate of 1e-5 (reduced by 40\% after every 30 epochs) and the model is trained for 100 epochs. The training takes around 8 minutes for 1 epoch for AFEW 8.0 training dataset with two NVIDIA Tesla K80 cards.

\subsection{Noisy Student Training \cite{xie2019self}} \label{sec: student}
\begin{figure}[t!]
\centering
\includegraphics[width=1.0\textwidth,height=29mm]{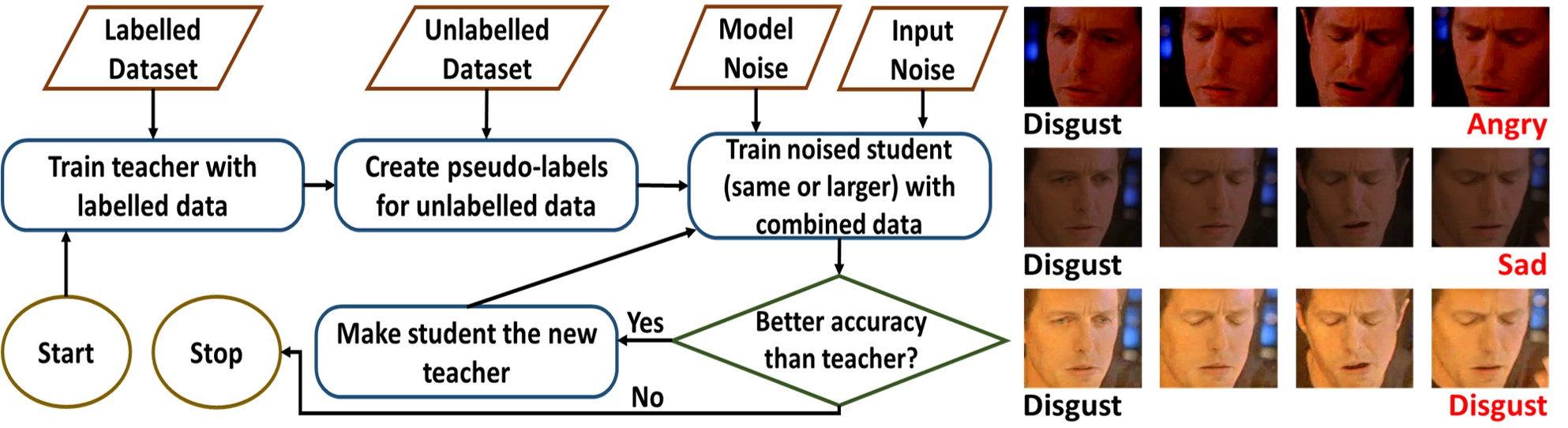}
\caption{Semi-supervised algorithm is presented in the flow-chart. We also show an example video from AFEW 8.0 dataset where the frames underwent different augmentations. Predictions without iterative training are shown in red and predictions after iterative training are shown in black.}
\label{fig:flowchart}
\end{figure}

Once the model is trained on the labelled set and the best possible model is obtained, we use it as a teacher model to create pseudo-labels on the subset of BoLD dataset that we collected. After generating the pseudo-labels, a student model (same size or larger than teacher) is trained on the combination of labelled and unlabelled dataset. While training the student model, we deliberately add noise in the form of random data augmentations and dropout (with 0.5 probability at the final hidden layer). Random data augmentations (using RandAugment \cite{cubuk2019randaugment}) include transformations such as brightness change, contrast change, translation, sharpness change and flips. RandAugment automatically applies $n\,\epsilon\,[2,4]$ random operations with a random magnitude $m\,\epsilon\,[0,9]$. After the noisy student is trained on the combined data, the trained student becomes the new teacher that generates new pseudo-labels for the unlabelled dataset. The iterative training continues until we observe a saturation in performance. From Fig. \ref{fig:flowchart}, we see how noisy training helps the student become more robust with the addition of noise. While the teacher may give different predictions for different alterations of the same video, the student is more accurate and stable with its predictions.

%

\section{Results}
In this section, we show the results obtained with and without iterative self-training, followed by comparison with state-of-the-art methods and ablation studies.
\begin{figure}[t!]
\centering
\includegraphics[width=1.0\textwidth,height=80mm]{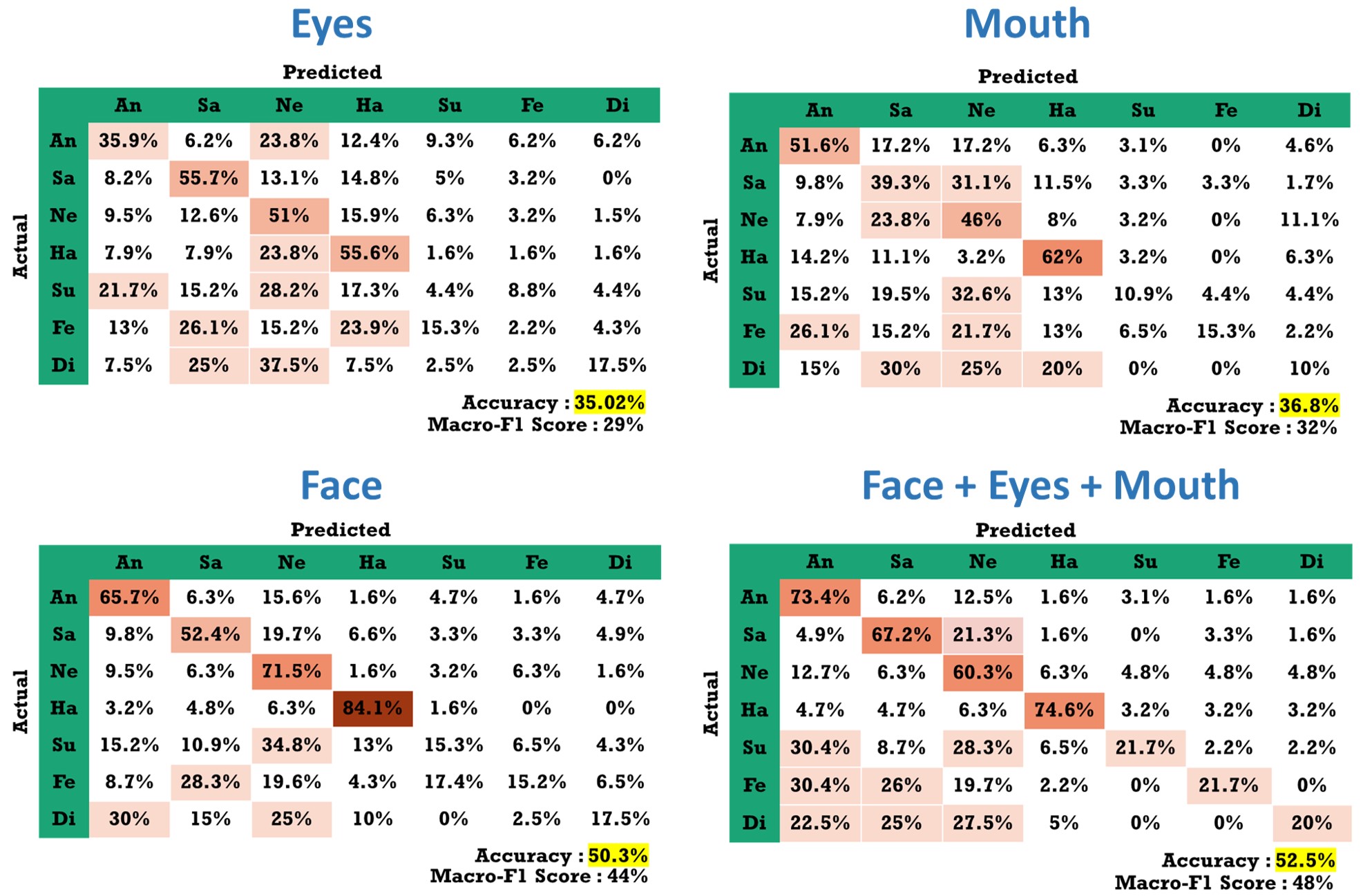}
\caption{This figure shows the confusion matrices, the accuracies, and the macro f1 scores achieved on the AFEW 8.0 dataset using different regions of the face. The proposed model (Face + Eyes + Mouth) achieves the highest accuracy. An=Angry, Sa=Sad, Ne=Neutral, Ha=Happy, Su=Surprise, Fe=Fear, Di=Disgust.}
\label{fig:confusion}
\end{figure}

\subsection{Without Student Training}
Fig. \ref{fig:confusion} shows the results of processing individual regions (without group convolution and channel attention) on the AFEW 8.0 dataset, along with the proposed methodology. Our objective is to explore a) if upper face region and lower face regions have different feedback signals that dominate different categories of emotions, and b) if isolating the regions and processing them independently leads to an increase of accuracy. As seen in the confusion matrix (Fig. \ref{fig:confusion}), the eyes region is better than the mouth region in the prediction of sadness and disgust categories. Intuitively, the squinted eyes expression in disgust and the droopy eyelids or furrowed eyebrows expression in sadness makes the eyes region pronounced. On the other hand, the mouth region is comparatively better with categories that require lip movements like happiness, anger, and surprise. Overall, 52.50\% accuracy is achieved using the proposed model, which is slightly better than the model that only uses faces. Furthermore, we see a significant increase in the macro f1 score when we include the eyes and mouth region along with faces indicating that the predictions are comparatively more unbiased for the seven categories (an advantage for noisy student training). The proposed model is still biased against fear, surprise, and disgust categories, but performs better than several existing methods \cite{lu2018multiple,acharya2018covariance,yan2018multi} where the reported accuracies for these categories are close to 0\%.

\subsection{With Iterative Training} \label{with_iterative}
Using noisy student training, we report our experimental results for four iterations on the AFEW 8.0 dataset and two iterations on the CK+ dataset.

\begin{figure}[t!]
\centering
\includegraphics[width=1.0\textwidth]{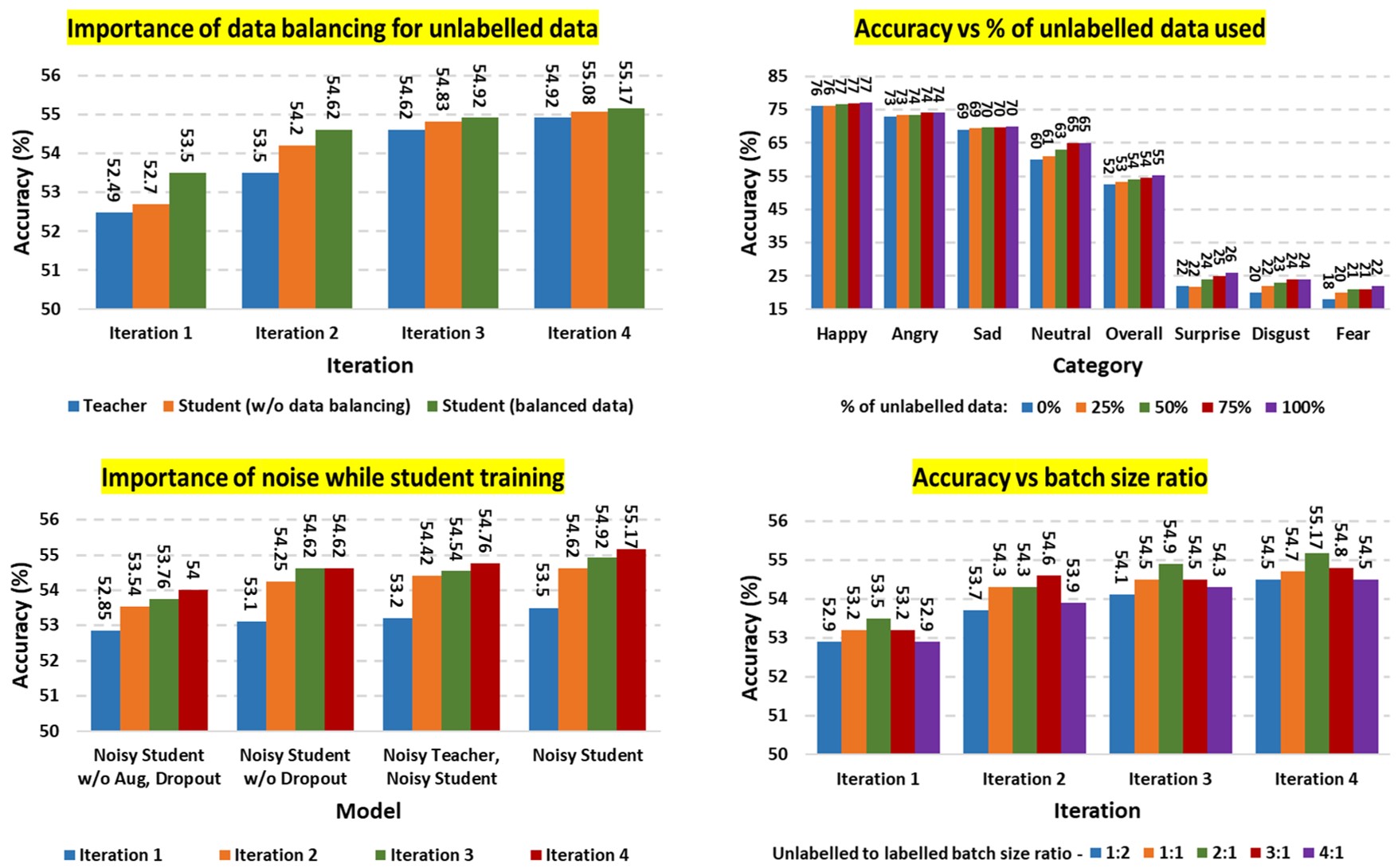}
\caption{This figure shows the experimental results of noisy student training for four iterations using AFEW 8.0 and BoLD datset.}
\label{fig:charts}
\end{figure}

\noindent\textbf{{Data Balancing:}}
Since the model is biased, the number of pseudo-labels in the unlabelled dataset for some categories is smaller than in other categories. We try to match the distribution of the training set by duplicating images of fear, disgust, and surprise categories. Additionally, images of angry, happy, and neutral classes are filtered out based on confidence scores. Fig. \ref{fig:charts} shows that balancing the pseudo-labels leads to better accuracy in each iteration compared to the student model without data balancing. The same trend is not observed for the CK+ dataset since the pseudo-labels roughly have the same distribution as the training set.

\noindent\textbf{{Unlabelled Dataset Size:}}
As stated in the original paper \cite{xie2019self}, using a large amount of unlabelled data leads to better accuracy. After data balancing, we use a fraction of the BoLD dataset and report the accuracy after several iterations of training until the performance saturates (see Fig. \ref{fig:charts}). For both CK+ and AFEW 8.0 dataset, we observe that using the whole unlabelled training set is better as opposed to using just a fraction of the dataset. Fig. \ref{fig:charts} shows a steady increase in all categories and overall accuracy with an increase in data size after four iterations of training on the AFEW 8.0 dataset.

\noindent\textbf{{Importance of Noise:}}
Noise helps the student to be more robust than the teacher, as addressed in Sec. \ref{sec: related}. The accuracy only reaches 53.5\% on the AFEW 8.0 dataset without noise in student training, and no improvement is seen on the CK+ dataset. However, we achieve an accuracy of 55.17\% after noisy training, which shows that input and model perturbations are vital while training the student. Additionally, Fig. \ref{fig:charts} shows that it is better when the pseudo-labels are generated without noise, i.e. the teacher remains as powerful as possible.

\noindent\textbf{{Batch Size Ratio:}}
When training on combined data, a batch of labelled images and a batch of unlabelled images are concatenated for each training step. If the batch sizes of labelled and unlabelled sets are equal, the model will complete several epochs of training on labelled data before completing one epoch of training on the BoLD dataset due to its larger size. To balance the number of epochs of training on both datasets, the batch size of the unlabelled set is kept higher than the labelled set. Fig. \ref{fig:flowchart} shows that a batch size ratio of 2:1 or 3:1 is ideal for training when AFEW 8.0 is used as the labelled training set. Similarly, a batch size ratio of 5:1 is ideal for the CK+ dataset.

\setlength{\tabcolsep}{4pt}
\begin{table}[t!] 
\begin{center}
\caption{We compare our results to the top-performing {\it single} models evaluated on the AFEW 8.0 dataset and state-of-the-art models evaluated on the CK+ dataset.}\label{tab: compare}

\begin{tabular}{cc|cc}
\toprule
\multicolumn{2}{c}{AFEW 8.0} & \multicolumn{2}{|c}{CK+} \\
\hline
\textbf{Models}& \textbf{Acc.} & \textbf{Models}& \textbf{Acc.}  \\
\hline
CNN-RNN (2016) \cite{fan2016video}& 45.43\%  & Lomo (2016) \cite{sikka2016lomo} & 92.00\%\\
DSN-HoloNet (2017) \cite{hu2017learning}& 46.47\%& CNN + Island Loss (2018) \cite{cai2018island}& 94.35\%\\
DSN-VGGFace (2018) \cite{fan2018video} & 48.04\% & 
FAN (2019) (Fusion) \cite{meng2019frame} & 94.80\%\\
VGG-Face + LSTM (2017) \cite{vielzeuf2017temporal}& 48.60\% & Hierarchial DNN (2019) \cite{kim2019efficient} & 96.46\%\\
VGG-Face (2019) \cite{aminbeidokhti2019emotion} & 49.00\% & DTAGN (2015) \cite{jung2015joint} & 97.25\%\\
ResNet-18 (2018) \cite{vielzeuf2018occam} & 49.70\% & MDSTFN (2019) \cite{sun2019deep} & 98.38\%\\
FAN (2019) \cite{meng2019frame} & 51.18\% & Compact CNN (2018) \cite{kuo2018compact} & 98.47\%\\
DenseNet-161 (2018) \cite{liu2018multi} & 51.44\% & ST Network (2017) \cite{zhang2017facial} & 98.47\%\\
Our Model (w/o iter. training) & 52.49\% & Our Model (w/o iter. learning) & 98.77\%\\
VGG-Face + BLSTM (2018) \cite{lu2018multiple} & 53.91\% & \textbf{FAN (2019) \cite{meng2019frame}} & \textbf{99.69\%}\\
\textbf{Our Model (iter. training)} & \textbf{55.17\%} & \textbf{Our Model (iter. learning)} & \textbf{99.69\%}\\

\bottomrule
\end{tabular}
\end{center}
\end{table}
\setlength{\tabcolsep}{1.4pt}

\subsection{Comparison with other methods} \label{sec: compare}
\begin{figure}[t!]
\centering
\includegraphics[width=1.0\textwidth, height=5cm]{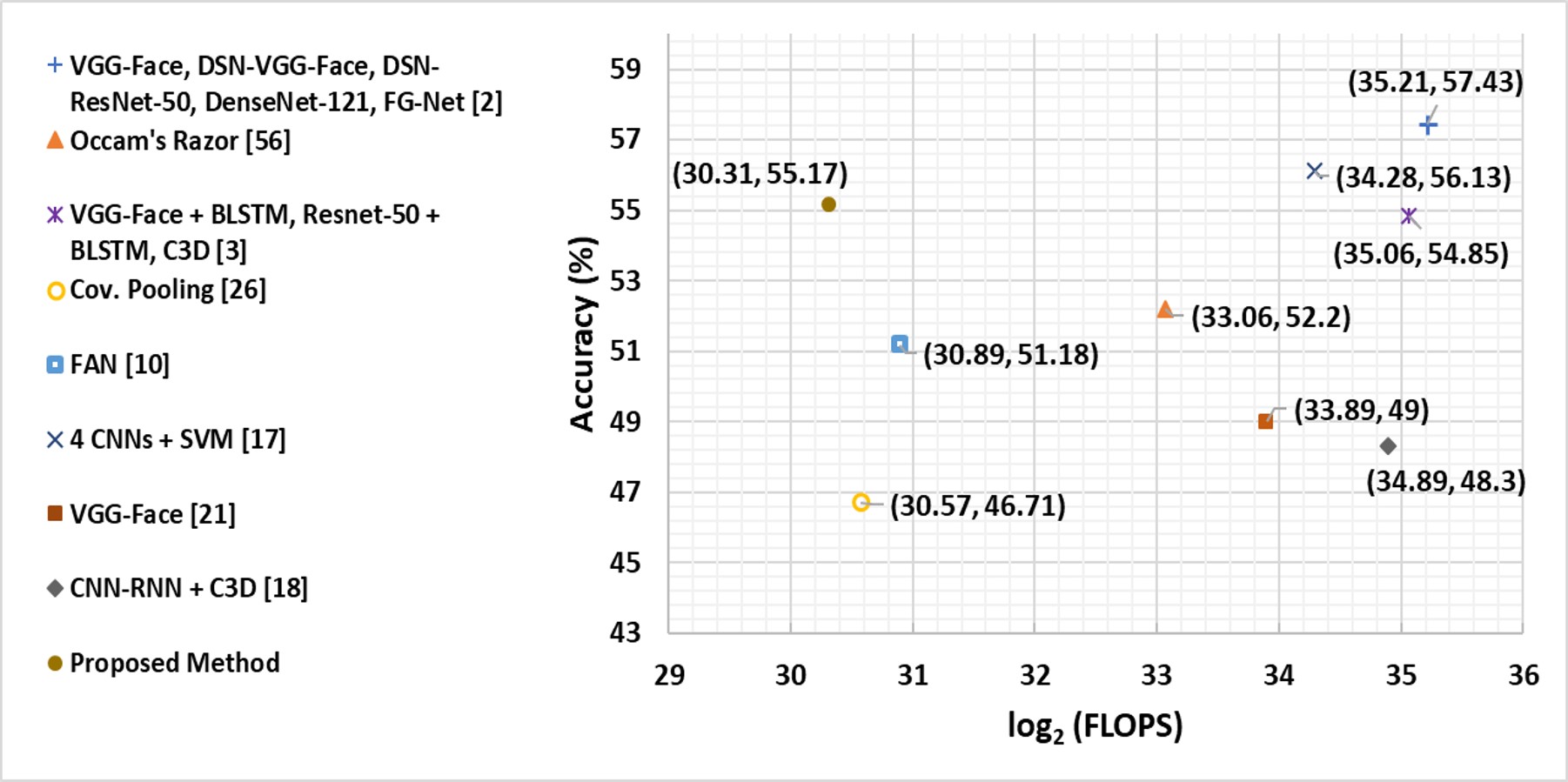}
\caption{Comparison of performance (in accuracy) vs computational cost (in FLOPS - Floating point operations per second) of state-of-the-art models evaluated on AFEW 8.0 dataset. FLOPS for the models are estimated values based on the backbone network unless explicitly specified by the authors. Most optimal models will be closer to the top-left corner.}
\label{fig:compare}
\end{figure}

We evaluate our model on the labelled datasets and show a comparison with the existing state-of-the art-methods (Table \ref{tab: compare}). On the AFEW 8.0 dataset, we achieve an accuracy of 52.5\% without iterative training and 55.17\% with iterative training. When comparing to existing best single models, our proposed method improves upon the current baseline \cite{lu2018multiple} by 1.6\%. Compared to static-based CNN methods that aim to combine frame scores for video-level recognition, we achieve a significant improvement of 3.73\% over the previous baseline \cite{liu2018multi}. We conduct a comparison of performance and speed of the existing state-of-the-art models including fusion methods (only visual modality) with our proposed model. Several methods that show higher validation accuracy have significantly higher computational demand which may be impractical for real-time world applications. For instance, \cite{vielzeuf2018occam} uses an ensemble of 50 models with the same architecture and yet attains a 52.2\% validation accuracy. Similarly, \cite{lu2018multiple,liu2018multi} use a combination of multiple deep learning models where each model has a higher computational cost than ours. We measure the computational complexity of state-of-the-art methods using FLOPS (Floating point operations) and results show that our method is the most optimal based on performance and speed (Fig. \ref{fig:compare}).

On the CK+ dataset, our method achieves an on par 10-fold cross-validation accuracy when compared to other state-of-the-art methods. While our model achieves an accuracy of only 98.77\% without iterative learning, the accuracy improves by 0.92\% when training data of each fold is combined with the unlabelled dataset for two iterations. This confirms our premise that self-training using noisy student is a robust procedure and can be used to increase the performance of a model on several other labelled data sources. Additionally, our results show that one can achieve better performance on a posed dataset when trained with an unlabelled in-the-wild dataset in a semi-supervised manner, which can be an effective alternative to labour-intensive tasks like gathering additional posed samples or labelling data.

\setlength{\tabcolsep}{4pt}
\begin{table}[t!] 
\begin{center}
\caption{This table shows the ablation studies conducted with AFEW 8.0 dataset. {\it Component Importance} shows the increase in accuracy with the addition of each component separately. {\it Noisy Student Training} shows the increase in accuracy with each loop of iterative learning and the effect of using a larger student.} \label{tab: ablation}
\label{table:headings}

 \begin{tabular}{cc|ccc}
\toprule
\multicolumn{2}{c}{\textbf{Component Importance}} & \multicolumn{3}{|c}{\textbf{Noisy Student Training}} \\
\hline
\textbf{Component}& \textbf{Acc.} & \textbf{Iteration} & \textbf{Student} & \textbf{Acc.}  \\
\hline
ResNet-18 (Baseline)  & 47.5\% & 0 &- &52.5\%\\
\cline{3-5}
+ MTCNN, Enlighten-GAN (Sec. \ref{sec: preprocess})& 48.3\% &\multirow{2}{*}{1}& ResNet-18 &53.5\%\\
+ Features from all blocks (Sec. \ref{sec: backbone}) & 49.3\% && ResNet-34 & 53.5\%\\
\cline{3-5}
+ Spatial-Attention (Sec. \ref{sec: spatial}) & 50.3\% &\multirow{2}{*}{2}& ResNet-18 &54.6\%\\
+ Multiple Regions (Sec. \ref{sec: backbone}) & 51.2\% && ResNet-34 &54.5\%\\
\cline{3-5}
+ Channel-Attention (Sec. \ref{sec: channel}) & 51.7\% &\multirow{2}{*}{3}& ResNet-18 &54.9\%\\
+ Frame-Attention (Sec. \ref{sec: frame}) & 52.5\% && ResNet-34 &54.8\%\\
\cline{3-5}
+ Iteration 1 - Self-training (Sec. \ref{sec: student}) & 53.5\% &\multirow{2}{*}{4}& ResNet-18 &55.2\%\\
+ Iteration 2 - Self-training (Sec. \ref{sec: student}) & 54.6\% && ResNet-34 &55.2\%\\
\cline{3-5}
+ Iteration 3 - Self-training (Sec. \ref{sec: student}) & 54.9\% &\multirow{2}{*}{5}& ResNet-18 &55.2\%\\
+ Iteration 4 - Self-training (Sec. \ref{sec: student}) & 55.2\%  && ResNet-34 &55.2\%\\
 
\bottomrule
\end{tabular}
\end{center}
\end{table}
\setlength{\tabcolsep}{1.4pt}

\subsection{Ablation Studies} \label{sec: ablation}
Our baseline model is ResNet-18 where the video-level feature vector is an unweighted average of all the frame-level feature vectors. Without sophisticated pre-processing, the baseline achieves an accuracy of 47.5\%. To better understand the significance of each component, we record our results after every change to the baseline model (Table \ref{tab: ablation}). Significant improvements are observed when features are concatenated from multiple residual blocks using spatial-attention, and when frame features are combined from multiple regions using group convolution and channel-attention.

Additionally, Table \ref{tab: ablation} shows the increase in validation accuracy with each loop of iterative learning. As suggested by \cite{xie2019self}, noisy student learning may perform better if the student is larger in size than the teacher.  Since ResNet-34 \cite{he2016deep} has a comparatively larger capacity, we report its results besides ResNet-18 as the student model for each iteration. As seen in Table \ref{tab: ablation}, our results do not show improvement when ResNet-18 in our student model is replaced with a larger backbone. A possible explanation is that the unlabelled dataset used by \cite{xie2019self} is a hundred times larger than the labelled dataset and using a student with higher capacity may have resulted in better performance. On the contrary, our unlabelled dataset is only four times larger than the labelled dataset. Gathering additional unlabelled samples and using a larger student may result in a further increase in accuracy on the AFEW 8.0 dataset.

\section{Conclusion}
We propose a multi-level attention model for video-based facial expression recognition, which is trained using a semi-supervised approach. Our contribution is a cost-effective single model that achieves on par performance with state-of-the-art models using two strategies. Firstly, we use attention with multiple sources of information to capture spatially and temporally important features, which is a computationally economical alternative to the fusion of multiple learning models. Secondly, we use self-training to overcome the lack of labelled video datasets for facial expression recognition. The proposed training scheme can be extended to other related tasks in the field of affective computing.

\section{Acknowledgements}
The authors acknowledge the support of Professor James Wang for providing the opportunity to work on this project during his course on Artificial Emotion Intelligence at the Pennsylvania State University.

\bibliographystyle{splncs}
\bibliography{0004}
\end{document}